\documentclass[runningheads]{llncs}
\usepackage[T1]{fontenc}
\usepackage{lmodern}
\usepackage[utf8]{inputenc}
\usepackage[letterpaper,pass]{geometry}
\usepackage{amsmath,amssymb,graphicx,booktabs,microtype,url}
\usepackage{tikz}
\usetikzlibrary{arrows.meta,positioning}
\usepackage{hyperref}
\hypersetup{hidelinks,pdftitle={CARET: Clinical Assertion-Aware Residual Calibration for Medical Image-Text Alignment},pdfauthor={Yifan Jin and Tao Liu}}
\newcommand{\norm}[1]{\left\lVert #1\right\rVert_2}
\newcommand{\ind}{\mathbb{I}}
\begin{document}
\title{CARET: Clinical Assertion-Aware Residual\\Calibration for Medical Image--Text Alignment}
\titlerunning{CARET for Medical Image--Text Alignment}
\author{Yifan Jin\inst{1} \and Tao Liu\inst{1}}
\authorrunning{Y. Jin and T. Liu}
\institute{Sun Yat-sen University, Guangzhou, Guangdong, China}
\maketitle
\begin{abstract}

Medical vision--language models align radiographs with free-text reports, but radiology language distinguishes explicit findings, negation, uncertainty, and omission. Treating these report states as interchangeable during feature adaptation can change image predictions without supporting visual evidence. We introduce CARET, a parameter-efficient image-side adaptation method that retains pretrained image and text encoders and uses paired positive and negative clinical statements to define a low-dimensional semantic correction space. A bounded residual adjusts image features along these directions, while assertion-aware supervision distinguishes explicitly present and absent findings from uncertain or unmentioned ones. At inference, the adapter requires only an image; textual references remain fixed. In zero-shot classification, CARET achieves 67.80\% accuracy on CheXpert 5$\times$200 and 65.20\% on RSNA. These scores exceed the published CLEFT-LoRA results by 1.06 and 0.27 percentage points, respectively. Comparisons to literature baselines are informative but not controlled for differences in pretraining and evaluation protocols. Ablations across eight components identify the four-state distinction and assertion objective as major contributors, with task-dependent effects for nonmention preservation and pair alignment.


\end{abstract}

\section{Introduction}

Chest radiographs are accompanied by reports documenting visual findings, uncertainty, and diagnostic impressions. Large-scale image--text pretraining has enabled cross-modal retrieval and zero-shot classification, with methods including ConVIRT, GLoRIA, MGCA, and PRIOR extending vision--language learning to clinical imaging \cite{zhang2022convirt,huang2021gloria,wang2022mgca,cheng2023prior}. These models commonly compare normalized image features with report or prompt embeddings. Consequently, the quality of their shared representation affects both retrieval and the ranking of diagnostic statements.

A difficulty specific to clinical reports is that meaning depends not only on which finding is mentioned but also on how it is asserted. For example, “pleural effusion,” “no pleural effusion,” and “cannot exclude a small pleural effusion” are not equivalent supervision signals. Nor is a report’s silence about a finding equivalent to a confirmed negative. This distinction underlies clinical uncertainty extraction in CheXpert \cite{irvin2019chexpert}, yet common global embedding objectives do not explicitly enforce the four-way distinction during post-pretraining adaptation.

The modality gap between image and text features offers a related but different perspective \cite{liang2022gap,an2025i0t}. Matching global feature statistics need not preserve positive--negative clinical distinctions. Instead, adaptation should correct image features along directions justified by report evidence, and avoid moving unsupported concepts simply because the paired report omitted them. Such a strategy is particularly suitable when pretrained encoders are strong and retraining them is prohibitively expensive.

We introduce CARET, an image-only adapter using directions defined by paired positive and explicitly negated medical prompts. Reports provide weak supervision during training; the adapted model has no dependency on reports or labels at inference. The update is confined to a clinically defined subspace and bounded to retain the underlying visual representation. Our formulation integrates explicit assertion agreement, assertion-local report guidance, and preservation of unmentioned concepts in a single lightweight objective.

We assess the resulting zero-shot recognition performance using CheXpert 5$\times$200 and RSNA and compare it with published medical vision--language and parameter-efficient adaptation baselines. Our contributions are an assertion-defined semantic edit space; a bounded one-sided update with separate treatment of negation, uncertainty, and nonmention; and a two-benchmark zero-shot classification study with component-wise ablations.

\section{Related Work}

\paragraph{Medical vision--language pretraining.}

ConVIRT learns visual features through paired radiograph--report contrastive training \cite{zhang2022convirt}. GLoRIA adds global and local alignment \cite{huang2021gloria}; MGCA uses multiple granularities \cite{wang2022mgca}; PRIOR combines prototype-based image--report learning \cite{cheng2023prior}; and MedCLIP addresses the limited availability of strictly paired examples \cite{wang2022medclip}. These models provide reusable medical image features and text references. CARET operates after pretraining instead of introducing another image--text backbone.

\paragraph{Visual evidence and clinical representation learning.}
Recent work has used report findings to guide radiograph segmentation
\cite{xi2026grounding} and shared radiographic representations for
clinical reasoning and evidence construction \cite{xi2026radiographic}.
Complementing these medical applications, HuLiRAG combines referent
detection, mask-based spatial grounding, and adaptive local--global
retrieval to guide multimodal answer generation \cite{xi2025hulirag}.
These approaches highlight the importance of grounding language in
visual evidence. CARET uses clinical assertion supervision to calibrate
frozen image embeddings for zero-shot recognition.
\paragraph{Efficient adaptation and modality alignment.}

CLEFT combines pretrained encoders with efficient language-side tuning and learned prompts and reports zero-shot results on CheXpert 5$\times$200 and RSNA \cite{du2024cleft}. I0T studies embedding standardization to reduce the modality gap \cite{an2025i0t}. CARET retains the text representation and adapts image features along positive--negative clinical prompt directions. Assertion-specific constraints control these semantic updates and preserve the scores of findings that a report does not address.

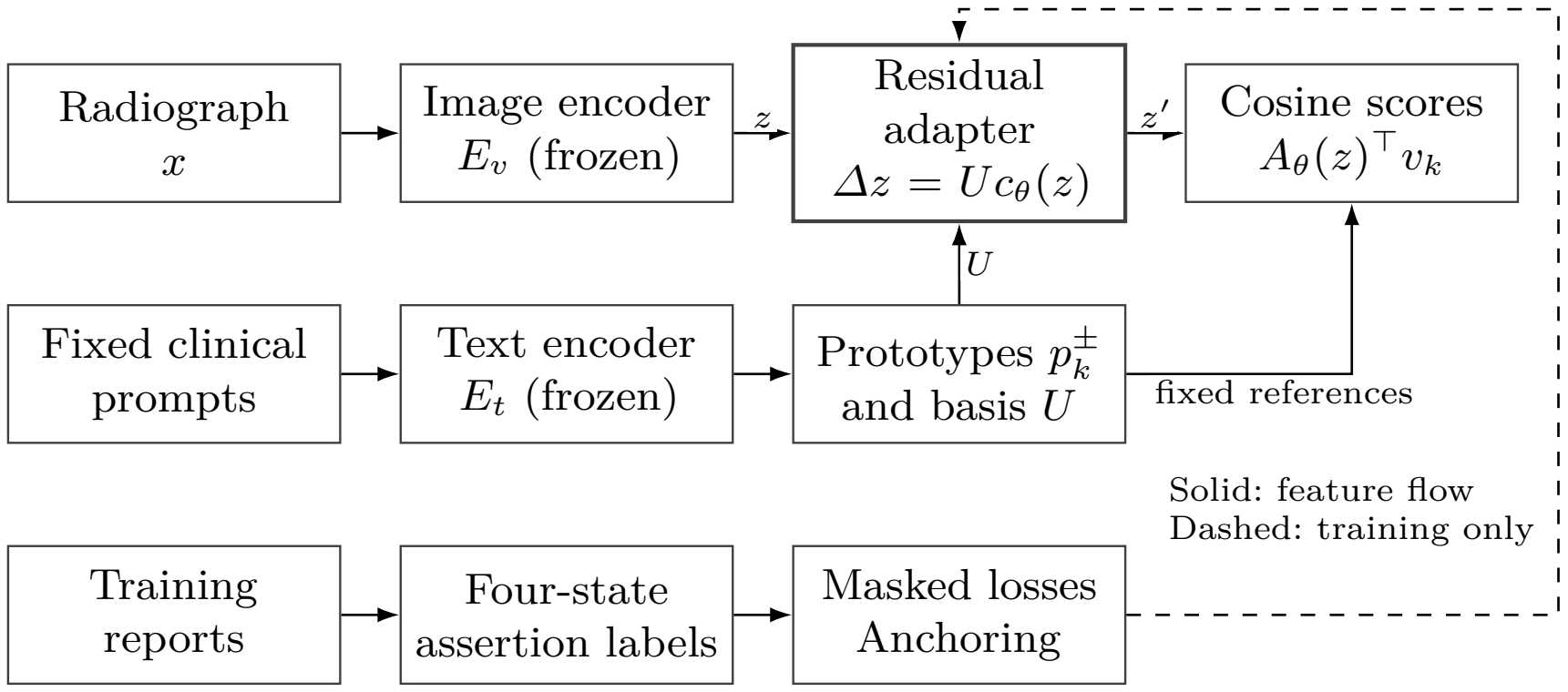
\begin{figure}[t]
\centering
\begin{tikzpicture}[
 x=1cm,y=1cm,
 box/.style={draw=black!75,line width=0.45pt,minimum height=10mm,
             text width=22mm,align=center,inner sep=3pt,font=\footnotesize},
 flow/.style={-{Latex[length=1.6mm,width=1.1mm]},line width=0.5pt},
 supervision/.style={flow,dashed},
 note/.style={font=\scriptsize,align=center,fill=white,inner sep=1pt}
]
\node[box] (image) at (0,0) {Radiograph\\$x$};
\node[box] (imageencoder) at (2.85,0) {Image encoder\\$E_v$ (frozen)};
\node[box,line width=0.8pt] (adapter) at (5.70,0)
 {Residual adapter\\$\Delta z=Uc_\theta(z)$};
\node[box] (prediction) at (8.55,0) {Cosine scores\\$A_\theta(z)^\top v_k$};

\node[box] (prompts) at (0,-1.75) {Fixed clinical\\prompts};
\node[box] (textencoder) at (2.85,-1.75) {Text encoder\\$E_t$ (frozen)};
\node[box] (basis) at (5.70,-1.75) {Prototypes $p_k^\pm$\\and basis $U$};

\node[box] (reports) at (0,-3.5) {Training\\reports};
\node[box] (assertions) at (2.85,-3.5) {Four-state\\assertion labels};
\node[box] (objectives) at (5.70,-3.5) {Masked losses\\Anchoring};

\draw[flow] (image)--(imageencoder);
\draw[flow] (imageencoder)--node[note,above] {$z$}(adapter);
\draw[flow] (adapter)--node[note,above] {$z'$}(prediction);
\draw[flow] (prompts)--(textencoder);
\draw[flow] (textencoder)--(basis);
\draw[flow] (basis.north)--node[note,right] {$U$}(adapter.south);
\draw[flow] (basis.east)-|node[note,pos=0.35,below] {fixed references}(prediction.south);
\draw[flow] (reports)--(assertions);
\draw[flow] (assertions)--(objectives);
\draw[supervision] (objectives.east)--(10.05,-3.5)--(10.05,0.9)
 --(5.70,0.9)--(adapter.north);
\node[font=\scriptsize,align=left] at (8.55,-2.75)
 {Solid: feature flow\\Dashed: training only};
\end{tikzpicture}
\caption{CARET framework. Frozen encoders provide image embeddings and fixed
clinical text prototypes. During training, reports supply assertion status to
supervise a bounded, image-side correction in the subspace $U$. During inference,
only a new image is needed; text prompts remain fixed.}
\label{fig:framework}
\end{figure}
\section{Method}
\subsection{Setting and Clinical Semantic Directions}
Let $\{(x_i,r_i)\}_{i=1}^{N}$ be paired radiographs and reports, and let
$E_v$ and $E_t$ denote pretrained visual and language encoders, including their
projection layers. We freeze both encoders and obtain normalized features
\begin{equation}
z_i=\frac{E_v(x_i)}{\norm{E_v(x_i)}},\qquad
b_i=\frac{E_t(r_i)}{\norm{E_t(r_i)}},\qquad z_i,b_i\in\mathbb{S}^{d-1}.
\end{equation}
For a vocabulary of $K$ radiographic concepts, we define positive prompt
templates $\mathcal{T}_k^+$ and explicitly negative templates $\mathcal{T}_k^-$.
Their text prototypes and corresponding semantic axis are
\begin{equation}
p_k^\pm=
\frac{\frac{1}{|\mathcal{T}_k^\pm|}\sum_{t\in\mathcal{T}_k^\pm}
\frac{E_t(t)}{\norm{E_t(t)}}}
{\norm{\frac{1}{|\mathcal{T}_k^\pm|}\sum_{t\in\mathcal{T}_k^\pm}
\frac{E_t(t)}{\norm{E_t(t)}}}},\quad
q_k=\frac{p_k^+-p_k^-}{\norm{p_k^+-p_k^-}}.
\end{equation}
Let $Q=[q_1,\ldots,q_K]$. A thin singular-value decomposition yields an
orthonormal basis $U\in\mathbb{R}^{d\times r}$ for a retained rank $r\leq K$.
This basis is computed once and fixed. It restricts changes to directions
identified by clinical statements, rather than allowing an unconstrained
update across the complete embedding space.

\subsection{Bounded Image-Side Residual Adapter}
Given $z\in\mathbb{S}^{d-1}$, a small MLP
$f_\theta:\mathbb{R}^d\rightarrow\mathbb{R}^r$ predicts correction coefficients
and a sigmoid gate $g_\theta(z)\in(0,1)$ controls their magnitude.
For a budget $0<\rho<1$, define
\begin{equation}
h_\theta(z)=g_\theta(z)\tanh(f_\theta(z)),\quad
c_\theta(z)=\rho\frac{h_\theta(z)}{\max\{1,\norm{h_\theta(z)}\}},
\end{equation}
\begin{equation}
A_\theta(z)=\frac{z+Uc_\theta(z)}{\norm{z+Uc_\theta(z)}}.
\end{equation}
The unnormalized residual lies in $\operatorname{span}(U)$ and has norm
at most $\rho$. Since $\norm{z}=1$, the normalized change satisfies
\begin{equation}
\norm{A_\theta(z)-z}\leq 2\rho.
\end{equation}
The bound limits the maximum perturbation of any cosine score against a unit
text vector to $2\rho$; it does not, by itself, guarantee an unchanged class
ranking. Only the adapter parameters $\theta$ are updated. Cached image and
report embeddings can be used throughout training.

\subsection{Four-State Assertion Supervision}
An assertion extractor maps report $r_i$ and concept $k$ to
$y_{ik}\in\{+1,-1,\mathrm{U},\mathrm{NM}\}$, where the four states denote
explicitly present, explicitly absent, uncertain, and not mentioned.
We define $m_{ik}=\ind[y_{ik}\in\{+1,-1\}]$ and
$u_{ik}=\ind[y_{ik}=\mathrm{NM}]$. Uncertain concepts contribute no signed
supervision, while unmentioned concepts receive only a preservation constraint.
Given a normalized vector $a$, a signed clinical score is
\begin{equation}
s_k(a)=\frac{a^\top(p_k^+-p_k^-)}{\tau},\qquad \tau>0.
\end{equation}
A masked logistic objective favors the correct polarity only for explicitly
asserted concepts:
\begin{equation}
\mathcal{L}_{\mathrm{assert}}=
\frac{\sum_{i,k}m_{ik}\operatorname{softplus}(-y_{ik}s_k(A_\theta(z_i)))}
{\sum_{i,k}m_{ik}+\epsilon}.
\end{equation}
The multiplication by $y_{ik}$ is evaluated only when $m_{ik}=1$.
To exploit the paired report without forcing entire feature vectors to
coincide, we add assertion-local guidance,
\begin{equation}
\mathcal{L}_{\mathrm{local}}=
\frac{\sum_{i,k}m_{ik}[q_k^\top(A_\theta(z_i)-b_i)]^2}
{\sum_{i,k}m_{ik}+\epsilon}.
\end{equation}
Importantly, an unmentioned finding is not treated as absent. Instead, we
preserve its original signed visual score:
\begin{equation}
\mathcal{L}_{\mathrm{NM}}=
\frac{\sum_{i,k}u_{ik}[s_k(A_\theta(z_i))-s_k(z_i)]^2}
{\sum_{i,k}u_{ik}+\epsilon}.
\end{equation}
This term anchors unmentioned concepts to their pretrained visual scores
without assuming that those scores are ground truth. A weak image--report similarity term
$\mathcal{L}_{\mathrm{pair}}=N^{-1}\sum_i[1-A_\theta(z_i)^\top b_i]$
maintains a link to retrieval, and
$\mathcal{L}_{\mathrm{reg}}=N^{-1}\sum_i\norm{c_\theta(z_i)}^2$
discourages unnecessary edits. The combined training criterion is
\begin{equation}
\begin{split}
\mathcal{L}={}&\mathcal{L}_{\mathrm{assert}}+
\lambda_l\mathcal{L}_{\mathrm{local}}+\lambda_n\mathcal{L}_{\mathrm{NM}}\\
&+\lambda_p\mathcal{L}_{\mathrm{pair}}+\lambda_r\mathcal{L}_{\mathrm{reg}},
\end{split}
\end{equation}
with nonnegative weights selected on validation data. Unlike batch-contrastive
training, this objective does not require unpaired samples to be designated
as negatives.
\subsection{Image-Only Inference}
For zero-shot recognition, each class is represented by a fixed ensemble of
encoded text prompts $v_k$. A test image is scored by $A_\theta(z)^\top v_k$
without accessing its report or an assertion label. For retrieval, the same
adapted image embedding can be compared with cached report embeddings; no
reverse adapter is introduced. Thus the learning signal is clinically informed
while inference remains image-only, and the original text representation is
preserved for reuse across tasks.
\section{Experiments and Results}
\subsection{Benchmarks and Evaluation}

We report zero-shot classification accuracy on two established chest-radiograph benchmarks. CheXpert 5$\times$200 contains five conditions with 200 images per condition and is commonly used to evaluate multi-class recognition \cite{irvin2019chexpert,huang2021gloria}. The RSNA pneumonia benchmark is used for cross-dataset classification assessment \cite{shih2019rsna}. In both settings, a radiograph is compared with candidate text prompts without fitting a downstream supervised classifier on the benchmark labels. The primary reference is the MICCAI 2024 CLEFT comparison \cite{du2024cleft}, which tabulates CLIP, ConVIRT, GLoRIA, MGCA, MedCLIP and efficient fine-tuning variants. We reproduce the zero-shot accuracy entries of its Table 1, together with CARET’s measured results. Published rows can differ in image encoder, text encoder, pretraining corpus, and evaluation details; the table is a literature comparison rather than a controlled rerun. No cross-method standard deviations are available for this set of values.

\subsection{Pretrained Encoders and Adaptation Data}
CLEFT-aligned DINOv2 and \mbox{BioMedLM-3B} provide the frozen image
and text branches, respectively \cite{du2024cleft}. The aligned
projection heads produce 512-dimensional embeddings. Matching
encoder families does not make the published CLEFT comparison
a controlled rerun.

The initial adaptation pool comprises the 222,758 official MIMIC-CXR
training studies, paired with reports and linked to the MIMIC-CXR-JPG
structured labels \cite{johnson2019mimiccxrjpg}.
Table~\ref{tab:data_inventory} lists the source inventory and official
partitions. The 1,808 validation studies and 3,269 test studies remain
separate from the training pool. Study counts are not image--report
pair counts because one study can contain multiple radiographs.
The final post-filtering image--report count is not reported separately.

\begin{table}[t]
\caption{MIMIC-CXR inventory. The first row counts images; all other rows count studies. The training pool precedes filtering.}
\label{tab:data_inventory}
\centering\small
\begin{tabular}{lr}
\toprule
Counting unit & Number\\
\midrule
Chest radiographs in the source database & 377,110\\
Imaging studies with corresponding reports & 227,835\\
Studies with CheXpert structured labels & 227,827\\
Official training studies: initial pool & 222,758\\
Official validation studies & 1,808\\
Official test studies & 3,269\\
\bottomrule
\end{tabular}
\end{table}

\subsection{Assertion Labels and Split Integrity}
Assertion supervision uses the CheXpert report-labeling schema
\cite{irvin2019chexpert,johnson2019mimiccxrjpg}. We retain the 13
concepts other than the aggregate \emph{No Finding} category:
atelectasis, cardiomegaly, consolidation, edema, enlarged
cardiomediastinum, fracture, lung lesion, lung opacity, pleural
effusion, pleural other, pneumonia, pneumothorax, and support devices.
The four-state mapping preserves the distinction between explicit
absence, uncertainty, and nonmention:
\[
 y_{ik}=
 \begin{cases}
 +1 & \text{for a positive report label }(1.0),\\
 -1 & \text{for an explicitly negative label }(0.0),\\
 \mathrm{U} & \text{for an uncertain label }(-1.0),\\
 \mathrm{NM} & \text{for a blank label cell.}
 \end{cases}
\]
The dataset code $-1.0$ denotes uncertainty, while the internal
target $-1$ denotes explicit absence. Blank cells in valid label rows
map to nonmention; a missing row does not. Explicit positive and
negative labels supervise assertion polarity, uncertain labels are
masked, and nonmention labels activate preservation.

Patient-level partitioning keeps all images of a patient together.
Study-level deduplication removes repeated study--report entries.
Development records are cross-checked against evaluation manifests.
The reported summary does not provide duplicate-match counts or
establish whether evaluation images appeared during encoder pretraining.

\subsection{Implementation and Optimization}
The shared embedding dimension is $d=512$, with $K=13$ clinical
concepts and a retained semantic-subspace rank of $r=8$.
The coefficient network has layer widths $512\rightarrow128\rightarrow8$,
and the adaptive gate has widths $512\rightarrow64\rightarrow1$.
The residual budget is $\rho=0.10$ and the clinical-score temperature
is $\tau=0.07$. Both pretrained encoders and the semantic basis remain
fixed; optimization updates the adapter parameters.

Table~\ref{tab:configuration} summarizes the training configuration.
AdamW uses a learning rate of $3\times10^{-4}$ and weight decay of
$10^{-4}$, with batches of 128 and a maximum of 20 epochs.
Early stopping has a patience of three epochs. The configuration
uses three random seeds. The accuracy tables present point estimates
and do not quantify run-to-run variability.

\begin{table}[t]
\caption{CARET architecture and optimization configuration.}
\label{tab:configuration}
\centering\small
\begin{tabular}{ll}
\toprule
Parameter & Value\\
\midrule
Shared feature dimension $d$ & 512\\
Clinical concepts $K$ & 13\\
Retained SVD rank $r$ & 8\\
Coefficient MLP & $512\rightarrow128\rightarrow8$\\
Adaptive gate & $512\rightarrow64\rightarrow1$\\
Residual budget $\rho$ & 0.10\\
Temperature $\tau$ & 0.07\\
Optimizer & AdamW\\
Learning rate & $3\times10^{-4}$\\
Weight decay & $10^{-4}$\\
Batch size & 128\\
Maximum epochs & 20\\
Early-stopping patience & 3 epochs\\
Number of random seeds & 3\\
\bottomrule
\end{tabular}
\end{table}

\subsection{Comparison with Published Models}

Table~\ref{tab:comparison} and Fig.~\ref{fig:accuracy} show that CARET reaches 67.80\% on CheXpert 5$\times$200, exceeding the 66.74\% reported for CLEFT-LoRA by 1.06 percentage points. This is the highest CheXpert accuracy among the methods included in the literature comparison. CLEFT-IA$^3$ and CLEFT-Prefix report 65.61\% and 60.55\%, respectively. Models without CLEFT’s efficient language-model tuning achieve lower CheXpert accuracies in the same assembled reference table, although their training regimes are heterogeneous.

\begin{table}[t]
\caption{Zero-shot classification accuracy (\%) on CheXpert 5$\times$200 and
RSNA. Literature results are reproduced from CLEFT \cite{du2024cleft},
Table~1. Higher is better. Differences in pretraining and evaluation settings
should be considered when interpreting cross-paper comparisons.}
\label{tab:comparison}
\centering
\begin{tabular}{lrr}
\toprule
Method & CheXpert 5$\times$200 & RSNA\\
\midrule
CLIP-ViT & 20.16 & 49.89\\
ConVIRT-R50 & 41.88 & 47.31\\
GLoRIA-R50 & 43.28 & 33.06\\
MGCA-ViT & 51.35 & 69.90\\
MedCLIP-Swin & 59.42 & 74.47\\
\midrule
CLEFT-Prefix & 60.55 & 64.07\\
CLEFT-IA$^3$ & 65.61 & 64.08\\
CLEFT-LoRA & 66.74 & 64.93\\
\midrule
\textbf{CARET} & \textbf{67.80} & \textbf{65.20}\\
\bottomrule
\end{tabular}
\end{table}
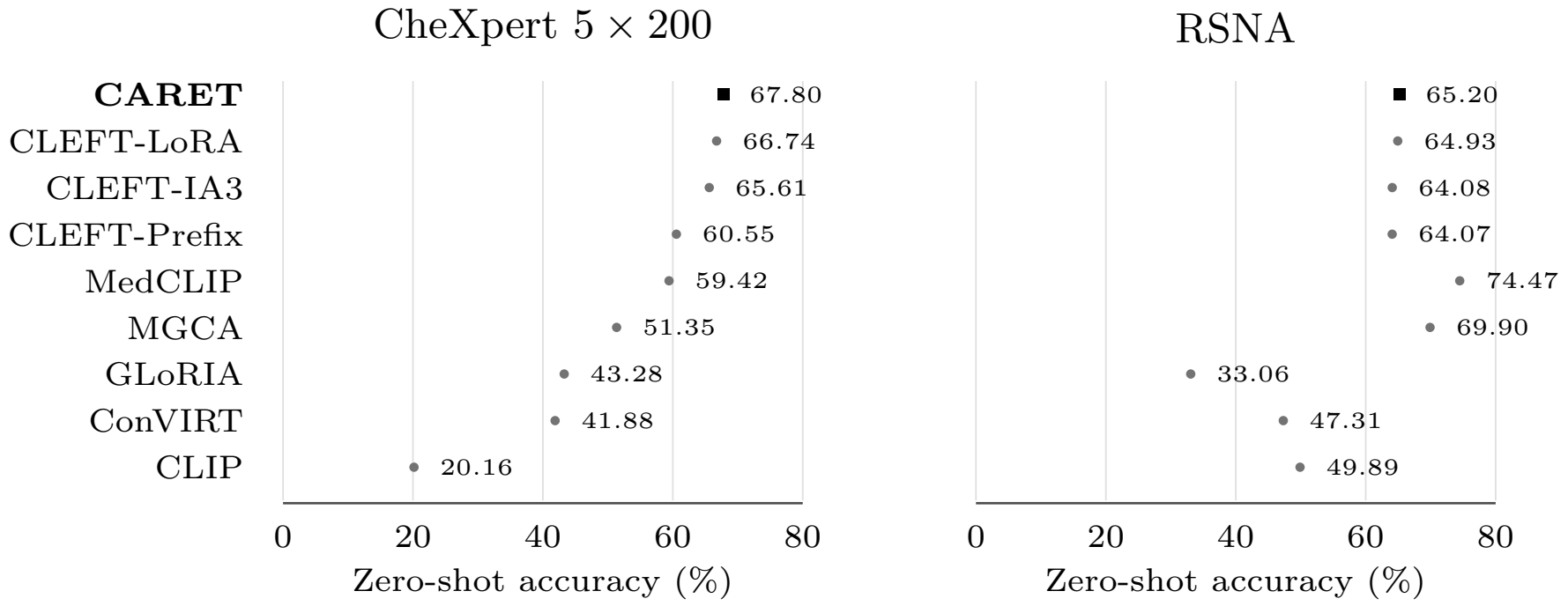
\begin{figure}[t]
\centering
\begin{tikzpicture}[x=1cm,y=1cm,font=\scriptsize]
\foreach \tick in {0,20,40,60,80}{
  \pgfmathsetmacro{\xa}{1.9+0.04875*\tick}
  \pgfmathsetmacro{\xb}{7.1+0.04875*\tick}
  \draw[black!12,line width=0.35pt] (\xa,0.1)--(\xa,3.25);
  \draw[black!12,line width=0.35pt] (\xb,0.1)--(\xb,3.25);
  \node[below] at (\xa,0.02) {\tick};
  \node[below] at (\xb,0.02) {\tick};
}
\draw[black!70,line width=0.5pt] (1.9,0.08)--(5.8,0.08);
\draw[black!70,line width=0.5pt] (7.1,0.08)--(11.0,0.08);
\node[font=\small] at (3.85,3.65) {CheXpert $5\times200$};
\node[font=\small] at (9.05,3.65) {RSNA};
\node at (3.85,-0.52) {Zero-shot accuracy (\%)};
\node at (9.05,-0.52) {Zero-shot accuracy (\%)};
\foreach \method/\a/\b/\row in {
 CLIP/20.16/49.89/1,
 ConVIRT/41.88/47.31/2,
 GLoRIA/43.28/33.06/3,
 MGCA/51.35/69.90/4,
 MedCLIP/59.42/74.47/5,
 CLEFT-Prefix/60.55/64.07/6,
 CLEFT-IA3/65.61/64.08/7,
 CLEFT-LoRA/66.74/64.93/8,
 CARET/67.80/65.20/9
}{
 \pgfmathsetmacro{\yy}{0.35*\row}
 \pgfmathsetmacro{\xa}{1.9+0.04875*\a}
 \pgfmathsetmacro{\xb}{7.1+0.04875*\b}
 \ifnum\row=9
   \node[anchor=east,font=\scriptsize\bfseries] at (1.72,\yy) {\method};
   \fill[black] (\xa-0.045,\yy-0.045) rectangle (\xa+0.045,\yy+0.045);
   \fill[black] (\xb-0.045,\yy-0.045) rectangle (\xb+0.045,\yy+0.045);
 \else
   \node[anchor=east] at (1.72,\yy) {\method};
   \fill[black!55] (\xa,\yy) circle[radius=0.035];
   \fill[black!55] (\xb,\yy) circle[radius=0.035];
 \fi
 \node[anchor=west,font=\tiny] at (\xa+0.08,\yy) {\a};
 \node[anchor=west,font=\tiny] at (\xb+0.08,\yy) {\b};
}
\end{tikzpicture}
\caption{Zero-shot accuracy on CheXpert 5$\times$200 and RSNA. CARET is
shown with black squares; gray circles denote the published comparisons
in CLEFT \cite{du2024cleft}. The horizontal axes start at zero to retain
the scale of differences.}
\label{fig:accuracy}
\end{figure}

On RSNA, CARET reaches 65.20\%, or 0.27 percentage points above CLEFT-LoRA’s Table 1 score of 64.93\%. MGCA-ViT and MedCLIP-Swin report 69.90\% and 74.47\%, respectively, and therefore remain stronger on this benchmark within the same literature comparison. These findings are more consistent with a setting-dependent benefit than with uniform superiority across all zero-shot tasks. Without matched runs and uncertainty estimates, the observed differences cannot be assigned statistical significance.

\subsection{Component Ablation}

Table~\ref{tab:ablation} summarizes eight leave-one-component-out variants under the reported evaluation protocol. Each value is zero-shot classification accuracy (\%) and the $\Delta$ columns give changes in percentage points relative to full CARET. The ablations evaluate supervision terms (assertion, local, pair, nonmention, residual), the semantic subspace and adaptive gate, and the explicit four-state assertion formulation.

\begin{table}[t]
\caption{Component ablation of CARET on CheXpert 5$\times$200 and RSNA
(accuracy, \%). $\Delta$ indicates the accuracy difference relative to the
full model in percentage points; negative values indicate degradation.}
\label{tab:ablation}
\centering\small
\begin{tabular}{lrrrr}
\toprule
Variant & CheXpert & $\Delta$ & RSNA & $\Delta$\\
\midrule
\textbf{Full CARET} & \textbf{67.80} & -- & \textbf{65.20} & --\\
\midrule
w/o Assertion Loss & 65.30 & $-2.50$ & 63.40 & $-1.80$\\
w/o Local Alignment & 67.00 & $-0.80$ & 64.50 & $-0.70$\\
w/o Nonmention Preservation & 68.00 & $+0.20$ & 64.40 & $-0.80$\\
w/o Pair Alignment & 67.40 & $-0.40$ & 65.30 & $+0.10$\\
w/o Residual Regularization & 67.20 & $-0.60$ & 64.70 & $-0.50$\\
w/o Clinical Subspace & 67.70 & $-0.10$ & 63.80 & $-1.40$\\
w/o Adaptive Gate & 67.10 & $-0.70$ & 64.60 & $-0.60$\\
w/o Four-State Distinction & 64.80 & $-3.00$ & 62.10 & $-3.10$\\
\bottomrule
\end{tabular}
\end{table}

Removing four-state distinction causes the largest drops (3.00 points on CheXpert and 3.10 on RSNA); removing assertion loss decreases accuracy by 2.50 and 1.80 points. Removing the clinical subspace mainly affects RSNA (-1.40 points). Nonmention preservation and pair alignment produce different tradeoffs: without preservation CheXpert rises by 0.20 while RSNA falls by 0.80 points, whereas without pair alignment RSNA rises by 0.10 while CheXpert falls by 0.40 points.

\paragraph{Reproducibility and evaluation scope.}
The architecture, initial data pool, assertion mapping, split-integrity
procedure, and optimization settings describe the experimental design.
Exact reproduction additionally requires the aligned checkpoint revision,
label release, final filtered training and validation manifests, and
overlap-audit outputs. Source inventory counts do not replace these
run-specific records, and published baseline comparisons remain descriptive.

\section{Conclusion}

CARET applies clinically supervised residual calibration to frozen image embeddings while keeping inference image-only. It achieves 67.80\% zero-shot accuracy on CheXpert 5$\times$200 and 65.20\% on RSNA. Ablations identify four-state assertions as particularly consequential (-3.00 and -3.10 points without them), while other terms show benchmark-dependent effects.

\bibliographystyle{splncs04}

\bibliography{references}

@inproceedings{an2025i0t,
 author={An, N. M. and Kim, E. and Thorne, J. and Shim, H.},
 title={{I0T}: Embedding Standardization Method Towards Zero Modality Gap},
 booktitle={Proceedings of the 63rd Annual Meeting of the Association for Computational Linguistics},
 year={2025}
}

@inproceedings{cheng2023prior,
 author={Cheng, P. and Lin, L. and Lyu, J. and others},
 title={{PRIOR}: Prototype Representation Joint Learning from Medical Images and Reports},
 booktitle={Proceedings of the IEEE/CVF International Conference on Computer Vision},
 year={2023}
}

@inproceedings{du2024cleft,
 author={Du, Y. and Chang, B. and Dvornek, N. C.},
 title={{CLEFT}: Language-Image Contrastive Learning with Efficient Large Language Model and Prompt Fine-Tuning},
 booktitle={Medical Image Computing and Computer Assisted Intervention -- MICCAI 2024},
 volume={15012},
 pages={465--475},
 publisher={Springer},
 year={2024},
 doi={10.1007/978-3-031-72390-2_44}
}

@inproceedings{huang2021gloria,
 author={Huang, S. C. and Shen, L. and Lungren, M. P. and Yeung, S.},
 title={{GLoRIA}: A Multimodal Global-Local Representation Learning Framework for Label-Efficient Medical Image Recognition},
 booktitle={Proceedings of the IEEE/CVF International Conference on Computer Vision},
 year={2021}
}

@inproceedings{irvin2019chexpert,
 author={Irvin, J. and Rajpurkar, P. and Ko, M. and others},
 title={{CheXpert}: A Large Chest Radiograph Dataset with Uncertainty Labels and Expert Comparison},
 booktitle={Proceedings of the AAAI Conference on Artificial Intelligence},
 year={2019}
}

@inproceedings{liang2022gap,
 author={Liang, W. and Zhang, Y. and Kwon, Y. and Yeung, S. and Zou, J.},
 title={Mind the Gap: Understanding the Modality Gap in Multi-Modal Contrastive Representation Learning},
 booktitle={Advances in Neural Information Processing Systems},
 year={2022}
}

@article{shih2019rsna,
 author={Shih, G. and Wu, C. N. and Halabi, S. S. and others},
 title={Augmenting the National Institutes of Health Chest Radiograph Dataset with Expert Annotations of Possible Pneumonia},
 journal={Radiology: Artificial Intelligence},
 volume={1},
 number={1},
 pages={e180041},
 year={2019}
}

@inproceedings{wang2022mgca,
 author={Wang, F. and Zhou, Y. and Wang, S. and Vardhanabhuti, V. and Yu, L.},
 title={Multi-Granularity Cross-Modal Alignment for Generalized Medical Visual Representation Learning},
 booktitle={Advances in Neural Information Processing Systems},
 year={2022}
}

@inproceedings{wang2022medclip,
 author={Wang, Z. and Wu, Z. and Agarwal, D. and Sun, J.},
 title={{MedCLIP}: Contrastive Learning from Unpaired Medical Images and Text},
 booktitle={Proceedings of the 2022 Conference on Empirical Methods in Natural Language Processing},
 year={2022}
}

@article{xi2025hulirag,
 author={Xi, Suyang and Yang, Chenxi and Ding, Hong and Ni, Yiqing and Liu, Catherine C. and Liu, Yunhao and Zhang, Chengqi},
 title={Taming a Retrieval Framework to Read Images in Humanlike Manner for Augmenting Generation of {MLLMs}},
 journal={arXiv preprint arXiv:2510.10426},
 year={2025},
 doi={10.48550/arXiv.2510.10426},
 url={https://arxiv.org/abs/2510.10426}
}

@article{xi2026grounding,
 author={Xi, Suyang and Hu, Songtao and Wang, Shansong and Ding, Hong and Li, Yuheng and He, Wenfeng and Zhang, Kuo and Del Balzo, Luke and Zhong, Chong and Hu, Mingzhe and Liu, Catherine C. and Liu, Yunhao and Yang, Xiaofeng},
 title={Grounding Radiology Report Findings into Medical Image Segmentation},
 journal={npj Digital Medicine},
 year={2026},
 doi={10.1038/s41746-026-03051-0}
}

@article{xi2026radiographic,
 author={Xi, Suyang and Hu, Songtao and Wang, Shansong and Safari, Mojtaba and del Balzo, Luke and Ul Karim, Ehsan and Hu, Mingzhe and Zhang, Kuo and Wang, Tonghe and Weichselbaum, Ralph R. and Yang, Xiaofeng},
 title={A Radiographic World Model for Clinical Reasoning and Evidence Generation},
 journal={arXiv preprint arXiv:2609.07719},
 year={2026},
 url={https://arxiv.org/abs/2609.07719}
}

@inproceedings{zhang2022convirt,
 author={Zhang, Y. and Jiang, H. and Miura, Y. and others},
 title={Contrastive Learning of Medical Visual Representations from Paired Images and Text},
 booktitle={Proceedings of Machine Learning for Healthcare},
 year={2022}
}

@article{johnson2019mimiccxrjpg,
 author={Johnson, A. E. W. and Pollard, T. J. and Berkowitz, S. and Greenbaum, N. R. and Lungren, M. P. and Deng, C. Y. and Mark, R. G. and Horng, S.},
 title={{MIMIC-CXR-JPG}, a Large Publicly Available Database of Labeled Chest Radiographs},
 journal={arXiv preprint arXiv:1901.07042},
 year={2019},
 doi={10.48550/arXiv.1901.07042},
 url={https://arxiv.org/abs/1901.07042}
}

\end{document}